\documentclass{llncs}

\makeatletter
\newcommand{\@chapapp}{\relax}%
\makeatother

\usepackage[title,toc,titletoc,header]{appendix}

\usepackage{tabularx}
\usepackage{graphicx}
\usepackage{subfig}
\usepackage{multirow}
\usepackage{mathtools}
\usepackage{amssymb}
\usepackage{comment}
\usepackage{algorithmicx}
\usepackage{algpseudocode}
\usepackage{algorithm}
\usepackage{color}
\usepackage{url}
\usepackage[outdir=./]{epstopdf}
\graphicspath{ {./} }
\begin{document}

\title{Not all Embeddings are created Equal: Extracting Entity-specific Substructures for RDF Graph Embedding}\author{Muhammad Rizwan Saeed\inst{1} \and Charalampos Chelmis\inst{2} \and Viktor K. Prasanna\inst{1}}
\institute{Ming Hseih Department of Electrical Engineering, University of Southern California, USA \\ \email{\{saeedm,prasanna\}@usc.edu}
\and Department of Computer Science, University at Albany - SUNY, USA \\ \email{cchelmis@albany.edu}}

\maketitle
\begin{abstract}
Knowledge Graphs (KGs) are becoming essential to information systems that require access to structured data. Several approaches have been recently proposed, for obtaining vector representations of KGs suitable for Machine Learning tasks, based on identifying and extracting relevant graph substructures using uniform and biased random walks. However, such approaches lead to representations comprising mostly ``popular'', instead of ``relevant'', entities in the KG. In KGs, in which different types of entities often exist (such as in Linked Open Data), a given target entity may have its own distinct set of most ``relevant'' nodes and edges. We propose \textit{specificity} as an accurate measure of identifying most relevant, entity-specific, nodes and edges. We develop a scalable method based on bidirectional random walks to compute specificity. Our experimental evaluation results show that specificity-based biased random walks extract more ``meaningful'' (in terms of size and relevance) RDF substructures compared to the state-of-the-art and, the graph embedding learned from the extracted substructures, outperform existing techniques in the task of entity recommendation in DBpedia.
\end{abstract}
\vspace{0.3cm}
\section{Introduction}
\label{sec:introduction}
Knowledge Graphs (KGs) i.e., graph structured knowledge bases, store information as entities and the relationships between them, often following some schema or ontology \cite{berners2001semantic}. With the advent of Linked Open Data \cite{bizer2009linked}, DBpedia \cite{AtzoriD14}, and Google Knowledge Graph\footnote{\url{https://www.blog.google/products/search/introducing-knowledge-graph-things-not/}}, large-scale KGs have drawn a lot of attention and have become important data sources for many Artificial Intelligence (AI) and Machine Learning (ML) tasks \cite{nickel2016review,rettinger2012mining}. As AI and ML algorithms work with propositional representation of data (i.e., feature vectors) \cite{RistoskiP16}, several adaptations of language modeling approaches such as word2vec \cite{Mikabs-1301-3781,MikolovSCCD13} and GloVe \cite{PenningtonSM14} have been proposed for generating graph embedding for entities in a KG. 
As a first step for such approaches, a ``representative'' neighborhood for each target entity in the KG must be acquired. 
To accomplish this task, approaches based on biased random walks \cite{CochezRPP17,GroverL16} have been proposed. These approaches use weighting schemes to make certain edges or nodes more likely to be included in the extracted subgraphs than others. 
Weighting schemes based on metrics such as frequency or PageRank \cite{CochezRPP17,Thalhammer2016} tend to favor inclusion of ``popular'' (or densely connected) nodes in the ``representative'' subgraphs. This can sometimes lead to inclusion of semantically less relevant nodes and edges in the ``representative'' subgraphs of target entities \cite{SunHYYW11}. We assert that the ``representative'' neighborhoods for different types of entities (e.g., book, movie, athlete) in cross-domain KGs, such as DBpedia, may comprise distinct sets of characteristic relationships. Our objective is to automatically identify these relationships and use them to extract entity-specific representations. This is in contrast to the scenario where extracted representations are KG-specific because of inclusion of popular nodes and edges, irrespective of their semantic relevance to the target entities. Additionally, we want to identify the most relevant neighborhood of a target entity without venturing into ``unrelated'' neighborhoods of close-by entities. For example, when identifying most relevant representation for a film, the director's name should be more likely to be included in the identified representation than his year or place of birth.

To address this challenge, we propose \textit{specificity} as an accurate measure for assigning weights to those semantic relationships which constitute the most intuitive and interpretable representations for a given set or type of entities. 
We provide a scalable method of computing specificity for semantic relationships of any depth in large scale KGs. 
We show that specificity-based biased random walks can extract more compact entity representations as compared to the state-of-the-art. 
To further demonstrate the efficacy of our specificity-based approach, we train neural language models (Skip-Gram \cite{Mikabs-1301-3781}) for generating graph embedding from the extracted entity-specific representations and use the generated embedding for the information retrieval task of entity recommendation.

The rest of this paper is structured as follows. In Section \ref{sec:relatedwork}, we provide a brief overview of related work. In Section \ref{sec:specificity}, we provide the necessary background and then motivate and introduce the concept of specificity. In Section \ref{sec:bider}, we present a scalable method for computing specificity. In Section \ref{sec:evaluation}, we present results highlighting beneficial characteristics of specificity using DBpedia dataset. In Section \ref{sec:conclusion} we present the conclusion and possible directions for future work.
\vspace{0.3cm}
\section{Related Work}
\label{sec:relatedwork}
\textbf{Graph Embedding Techniques:}~Numerous techniques have been proposed for generating appropriate representations of KGs for AI and ML tasks. Graph kernel based approaches simultaneously transverse the neighborhoods of a pair of entities in the graph to compute kernel functions based on metrics such as number of common substructures (e.g., paths or trees) \cite{LoschBR12,VriesR15} or graphlets \cite{ShervashidzeVPMB09}. Neural language models such as word2vec \cite{Mikabs-1301-3781,MikolovSCCD13} and GloVe \cite{PenningtonSM14}, originally proposed for generating word embedding, have been adapted for KGs. Deep Graph Kernel \cite{YanardagV15} identifies graph substructures (graphlets) and uses neural language models to compute a similarity matrix between different identified substructures. For large scale KGs, embedding techniques based on random walks have been proposed. 
DeepWalk \cite{PerozziAS14} learns graph embedding for nodes in the graph using neural language models while generating truncated uniform random walks. 
node2vec \cite{GroverL16} is a more generic approach than DeepWalk and uses $2^{nd}$ order biased random walks for generating graph embedding, preserving roles and community memberships of nodes. RDF2Vec \cite{RistoskiP16}, an extension of DeepWalk and Deep Graph Kernel, uses BFS-based random walks for extracting subgraphs from RDF graphs, which are converted into feature vectors using word2vec \cite{Mikabs-1301-3781,MikolovSCCD13}. RDF2Vec has been shown to outperform graph kernel-based approaches in terms of scalability and suitability for ML tasks for large scale KGs, such as DBpedia. 
The problem with uniform (or unbiased) random walks is the lack of control over explored neighborhood. To address this, biased random walks based approaches \cite{AtzoriD14,CochezRPP17,GroverL16} have been proposed which use different weighting schemes for nodes and edges. The weights create the \textit{bias} by making certain nodes or edges more likely to be visited during random walks. 
Biased RDF2Vec \cite{CochezRPP17} uses frequency-, degree-, and PageRank-based metrics for weighting schemes. Our proposed approach is closer to RDF2Vec \cite{CochezRPP17,RistoskiP16} in terms of extracting entity representations and using neural language model for generating embedding. 
The main difference is that we use our proposed metric of \textit{specificity} as an edge- and path-weighting scheme for biased random walks for identifying most relevant substructures for extracting entity representations from KGs. 

\noindent \textbf{Semantic Similarity and Relatedness:}~
Semantic similarity and relatedness between two entities have been relatively well explored \cite{AggarwalAZB15,GabrilovichM07,LealRQ12,PaulRMKS16}. 
Searching for similar or related entities given a search entity is a common task in information retrieval. However, before developing such functionality, it is important to define the notion of entity similarity and the set of attributes that will be used for its computation. Semantic similarity and relatedness are often used interchangeably in literature \cite{AggarwalAZB15,PaulRMKS16}, where similarity between two entities is computed based on common paths between them. This definition allows computation of similarity between any two entities, including entities of different types. 
For this paper, we assert that entities of different types carry different semantic meanings, whereas our objective is to automatically identify semantic relationships that constitute the representative neighborhoods of entities of the same given type. Therefore, we limit the computation of similarity to be between two entities of same type.  
PathSim \cite{SunHYYW11} is one of the approaches proposed for searching for similar entities in heterogeneous information networks. This approach is based on user-defined meta paths (i.e., sequence of relationships between entities) connecting entities of the same type. In contrast, our objective is to automatically identify the most relevant paths using specificity.

\section{Specificity: An Intuitive Relevance Metric}
\label{sec:specificity}
In this section, we introduce and motivate the use of \textit{specificity} as a novel metric for quantifying relevance. 

\subsection{Preliminaries}
\label{ss:prelim}
An RDF graph is represented by a knowledge base of triples \cite{TzitzikasLZ12}. A triple consists of three parts: \textit{$<$subject (s), predicate (p), object (o)$>$}. 
\begin{definition}{\textbf{RDF Graphs:}}
Assuming that there is a set $U$ of Uniform Resource Identifiers (URIs), a set $B$ of blank nodes, a set $L$ of literals, a set $O$ of object properties, and a set $D$ of datatype properties, an RDF graph $G$ can be represented as a set of triples such that:
\begin{align}
G = &\{<s,p,o> |~s \in (U \cup B),~p \in (D \cup O), (o \in (U \cup B),~if~p \in O)\nonumber\\
&\wedge (o \in L,~if~p \in D), ((D \cup O) \subseteq U)\}\nonumber
\end{align}
\end{definition}
In this paper, we simply represent an RDF graph $G$ as $G=\{V,E\}$ such that $V \in (U \cup B \cup L)$ and $E \in (O \cup D)$, where $E$ is a set of directed labeled edges. 
\begin{definition}{\textbf{Semantic Relationship:}}
A semantic relationship in an RDF graph can be defined as $<s,P,o>$ where $P$ can be a single predicate or a path represented by successive predicates and intermediate nodes between $s$ and $o$. 
\end{definition}
For this paper, we define semantic relationship $P^d$ of depth or length $d$, as a template for a path (or a walk) in $G$, that comprises of $d$ successive predicates $p_1,p_2,\ldots,p_d$. Thus, $<s,P^d,o>$ represents all paths (or walks) between any two entities $s$ and $o$ that traverse through $d-1$ intermediate nodes, using the same $d$ successive predicates that constitute $P^d$.
\begin{definition}{\textbf{RDF Graph Walks:}}
\label{def:walkdefinition}
Given a graph $G=\{V,E\}$, a single graph walk of depth $d$ starting from a node $v_0 \in V$, comprise of a sequence of $d$ edges (predicates) and $d$ nodes (excluding $v_0$): $v_0 \rightarrow e_1 \rightarrow v_1 \rightarrow e_2 \rightarrow v_2\rightarrow~\ldots~\rightarrow e_d \rightarrow v_d$. 
\end{definition}
Random graph walks provide a scalable method of extracting entity representations from large scale KGs \cite{GroverL16}. Starting from a node $v_0 \in V$, in the first iteration, a set of randomly selected outgoing edges $E_1$ are explored to get a set of nodes $V_1$ at depth 1. In the second iteration, from every $v \in V_1$, outgoing edges are randomly selected for exploring next set of nodes at depth 2. This is repeated until a set of nodes at depth $d$ is explored. The generated random walks are the union of explored triples during each of the $d$ iterations.  
This simple scheme of random walks, defined here, resembles a randomized breadth-first search. In literature both breadth-first and depth-first search strategies and an interpolation between the two have been proposed for extracting entity representations from large scale KGs \cite{GroverL16,PerozziAS14,RistoskiP16}. 

\subsection{Specificity}
\label{ss:specificity}
Consider the following example:
\begin{example}
\label{ex:batman}
Starting from the entity \textit{Batman (1989)} in DBpedia, a random walk explores following semantic relationships\footnote{Descriptive names used for brevity instead of actual URIs.}:
\begin{align}
&Batman~(1989) \xrightarrow{\text{director}} Tim~Burton \xrightarrow{\text{knownFor}} Gothic~Films\nonumber\\
&Batman~(1989) \xrightarrow{\text{director}} Tim~Burton \xrightarrow{\text{subject}} 1958~births\nonumber\\
&Batman~(1989) \xrightarrow{\text{director}} Tim~Burton \xrightarrow{\text{birthPlace}} Burbank,~CA\nonumber
\end{align}
\end{example}
Our intuition suggests that the style of a director (represented by \textit{Gothic Films}) is more relevant to a film than his year and place of birth. Frequency-, degree-, or PageRank-based metrics of assigning relevance may assign higher scores to nodes representing broader categories or locations. For example, PageRank scores (non-normalized) computed for DBpedia entities \textit{Gothic Films, 1958-births,} and \textit{Burbank, CA} are 0.586402, 161.258, and 57.1176 respectively\footnote{\url{http://people.aifb.kit.edu/ath/#DBpedia_PageRank}}. PageRank-based biased random walks may include these popular nodes and exclude intuitively more relevant information related to the target entity. Our objective is to develop a metric that assigns higher score to relevant nodes and edges in such a way that the node \textit{Gothic Films} becomes more likely to be captured, for \textit{Batman (1989)}, than \textit{1958 births} and \textit{Burbank, CA}. This way, the proposed metric will capture our intuition behind identifying more relevant information in terms of its specificity to the target entity. To quantify this relevance based on specificity, we determine if \textit{Gothic Films} represents information that is ``specific'' to \textit{Batman (1989)}. We trace all paths of depth $d$ reaching \textit{Gothic Films} and compute the ratio of number of those paths that originate from \textit{Batman (1989)} to number of all traced paths. This gives specificity of \textit{Gothic Films} to \textit{Batman (1989)} as a score between 0.0-1.0. A specificity score of 1.0 means that all paths of depth $d$ reaching \textit{Gothic Films} have originated from \textit{Batman (1989)}. 
For $G=\{V,E\}$, this node-to-node specificity of a node $n_1$ to $n_2$, such that $n_1 \in V$, $n_2 \in V$ and $P^d$ being any arbitrary path, can be defined as: 
\begin{align}
&Specificity(n_1,n_2)=\frac{|<n_2,P^d,n_1> \in G|}{|<v,P^d,n_1> \in G : v \in V|}
\end{align}
For the objective of using specificity for extracting relevant subgraphs, instead of defining specificity as a metric of relevance between each pair of entities (or nodes), we make two simplifying assumptions. First, we assert that each class or type of entities (e.g. movies, books, athletes, politicians) has a distinct set of characteristic semantic relationships. This enables us to compute specificity as a metric of relevance of a node (\textit{Gothic Films}) to a class or type of entities (\textit{Film}), instead of each individual instance of that class (e.g. \textit{Batman (1989)}). Second, we measure specificity of a semantic relationship (\textit{director,knownFor}), instead of an entity (\textit{Gothic Films}), to the class of target entities. Here, we are making the assumption that if majority of the entities (nodes) reachable via a given semantic relationship represents entity-specific information, we consider that semantic relationship to be highly \textit{specific} to the given class of target entities. From our example, this means that instead of measuring specificity of \textit{Gothic Films} to \textit{Batman (1989)}, we measure specificity of semantic relationship \textit{director,knownFor} to the class or entity type \textit{Film}. Based on these assumptions, we can define specificity as:
\begin{definition}{\textbf{Specificity:}}
Given an RDF graph $G=\{V,E\}$, a semantic relationship $P^d$ of depth $d$, and a set $S \subseteq V$ of all entities of type $t$, let $V_{S,P^d}\subseteq V$ be the set of all nodes reachable from $S$ via $P^d$. We define the specificity of $P^d$ to $S$ as
\begin{align}
&Specificity(P^d,S)=\frac{1}{|V_{S,P^d}|} \sum_{k \in V_{S,P^d}} \frac{| <s,Q^d,k>\in G : s \in S|}{| <v,Q^d,k>\in G:v \in V|}
\label{eq:specificity}
\end{align}
\label{def:specificity}
\end{definition}
$Q^d$ represents any arbitrary semantic relationship of length $d$. All $s \in S$ have a common associated type $t$, which means that $\forall s \in S, \exists<s,rdf:type,t> \in G$. Therefore, henceforth, we use the term $Specificity(P^d,t)$ instead of $Specificity(P^d,S)$ for denoting specificity.

\section{Bidirectional Random Walks for Computing Specificity}
\label{sec:bider}
Computing Equation \ref{eq:specificity} requires accessing large parts of the knowledge graph. In this section, we present an approach that uses bidirectional random walks to compute specificity. To understand, consider an entity type $t$ and a semantic relationship $P^d$, for which we want to compute $Specificity(P^d,t)$. We start with a set $S$ containing a small number of randomly selected nodes of type $t$. From nodes in $S$, forward random walks via $P^d$ are performed to collect a set of nodes $V_{S,P^d}$ (ignoring intermediate nodes, for $d>1$). From nodes in set $V_{S,P^d}$, reverse random walks in $G$ (or forward random walks in $G^r=reverse(G)$) are performed using arbitrary paths of length $d$ to determine the probability of reaching any node of type $t$. Specificity is computed as number of times a reverse walk lands on a node of type $t$ divided by total number of walks. This idea is the basis for the algorithm presented next which, for a given type $t$ in $G$, builds a list of most relevant semantic relationships up to depth $d$ sorted by their specificity to $t$.

\subsection{Algorithm}
The function \textit{rankBySpecificity} in Algorithm \ref{alg:propSel} performs initialization of variables and builds a set of semantic relationships for which specificity is to be computed. $Q_{paths}$ and $Q_{spec}[]$ hold the set of semantic relationships, unsorted and sorted by specificity respectively. $Q_{spec}[]$ is initialized as an array of size $d$ to hold sorted semantic relationships for every depth up to $d$. $N_{paths}$ specify the size of $Q_{paths}$. $N_{walks}$ is the number of bidirectional walks performed for computing specificity for each semantic relationship in $Q_{paths}$. 
A set $S$ of randomly selected nodes of type $t$ is generated in line 3. For each $i^{th}$ iteration ($i \leq d$), a set of semantic relationships $Q_{paths}$ is selected in line 5. The function \textit{computeSpecificity}, in line 6, computes specificity for each semantic relationship in $Q_{paths}$ and returns results in $Q_{spec}[i]$. Each element of $Q_{spec}$ is an array of dictionaries. Each dictionary contains $key-value$ pairs sorted by $value$, where $key$ is the semantic relationship and $value$ is its specificity. For each $i^{th}$ iteration of \textit{for} in Algorithm \ref{alg:propSel}, $Q_{paths}$ can be populated from scratch with semantic relationships of depth $i$ by random sampling of outgoing paths from $S$. Alternatively, for iterations $i \geq 2$, $Q_{paths}$ can be populated by expanding from most specific semantic relationships in $Q_{spec}[i-1]$. In our implementation of the algorithm, we use $Q_{spec}[i-1]$ for populating $Q_{paths}$ in $i^{th}$ iteration. 

\renewcommand{\algorithmicrequire}{\textbf{Input:}}
\renewcommand{\algorithmicensure}{\textbf{Output:}}
\begin{algorithm}[!t]
\caption{$rankBySpecificity(G, d, t)$}
\label{alg:propSel}
\begin{algorithmic}[1]
\small
\Require {RDF graph $G = \{V,E\}$, $d$ is the maximum depth of semantic relationships to be considered, originating from entities of type $t$.}
\Ensure {Returns ranked list $Q_{spec}$[] of semantic relationships for $depths \leq d$, with a score of $0.0-1.0$}
\State {initialize $Q_{paths},Q_{spec}[]$ to null/empty}
\State {initialize $N_{paths}, N_{walks}$}
\State {$S \leftarrow $ Generate random nodes of type $t$}
\For {$i \gets 1, d$} 
\State {$Q_{paths} \gets selectPaths(G, S, i, N_{paths})$}
\State {$Q_{spec}[i] \gets computeSpecificity(G, Q_{paths}, S, t, i, N_{walks})$}
\EndFor
\State {\textbf{return} \texttt{$Q_{spec}[]$}}
\end{algorithmic}
\end{algorithm}

Algorithm \ref{alg:reverse} shows the function \textit{computeSpecificity} which computes specificity for a given set of semantic relationships in $Q$ ($Q_{paths}$ from Algorithm \ref{alg:propSel}). In lines 6 and 7, for each semantic relationship $q \in Q$, a node $s \in S$ is randomly selected to get a node $v$ reachable from $s$ via $q$ in $G$ (\textit{forward walk}). 
In line 8, using $v$ and by selecting an arbitrary path $q'$ of depth $d$, a node $v'$ is reached (\textit{reverse walk}\footnote{Reverse walk in $G$ or forward walk in $G^r$}).
If $t$ is one of the types associated with $v'$, variable \textit{count} is incremented in line 10. This process is repeated $N$ times for each $q$, where $N > |S|$. At line 13, $Specificity(q,t)$ is computed as $\frac{count}{N}$. Lines 4-13 are repeated until $Specificity(q,t)$ has been computed for each $q \in Q$. The return variable $L$ contains semantic relationships and their specificity scores as $key-value$ pairs.

\renewcommand{\algorithmicrequire}{\textbf{Input:}}
\renewcommand{\algorithmicensure}{\textbf{Output:}}
\begin{algorithm}[!t]
\caption{$computeSpecificity(G, Q, S, t, d, N)$}
\label{alg:reverse}
\begin{algorithmic}[1]
\small
\Require {RDF graph $G = \{V,E\}$, $S$ is a set of random entities with a common type $t$, $Q$ is a set of semantic relationships of length $d$ to be processed, $N$ is number of bidirectional walks to be performed for each semantic relationship in $Q$}
\Ensure {Returns list $L$ of semantic relationships sorted by specificity ($0.0-1.0$)}
\State {$G^r = reverseEdges(G)$}
\State {initialize $dictionary$ $L$}
\ForAll {$q \in Q$}
\State {$count \leftarrow 0$}
\Repeat
\State {$s \leftarrow$ randomly pick a node from $S$}
\State {$v \leftarrow$ randomly explore node from $s$ using path $q$ in $G$} 
\State {$v'\leftarrow$ randomly explore node from $v$ using any path $q'$ in $G^r$} 
\If { $\exists <v', rdf:type, t> \in G$ }
\State { $count \leftarrow count + 1 $} 
\EndIf
\Until {$N$ times}
\State {insert $(q, \frac{count}{N})$ in $L$}
\EndFor
\State {\textbf{return} $L$}
\end{algorithmic}
\end{algorithm}

\subsection{Biased RDF Graph Walks with Pruning}
\label{sec:prunedRW}
Once, list of most relevant semantic relationships based on specificity is generated, we use it to create specificity-based biased random walks for extracting representative subgraphs for target entities. In order to further improve subgraph extraction, we outline some pruning schemes for the biased random walks. 

\subsubsection{Non-repeating Starting Entity (NRSE)}\label{sssec:nrse} \hfill \break
We start with the scheme that has least restrictive criteria for inclusion of entities in the graph walk of depth $d$. Starting from $v_0$ in $G$ (Definition \ref{def:walkdefinition}), we generate random graph walks of depth $d$. If $v_0$ is observed at depth $k$ such that $k < d$, the graph walk is discarded. 

\subsubsection{Unique Entities (UE)}\label{sssec:ue} \hfill \break
This scheme is more restrictive than \textit{NRSE} and does not allow repetition of any node in a single graph walk i.e., all nodes traversed in a single graph walk must be unique, resulting in a path. 
  
\subsubsection{Non-repeating Starting Entity Type (NRST)}\label{sssec:nrst}  \hfill \break
Assume that $m_1$ and $m_2$ are two entities of same type in an RDF graph and are connected by a path of depth $k$. A graph walk of depth $d$ from $m_1$, such that $d > k$, may also traverse attributes specific to $m_2$. To avoid this, this pruning strategy discards graph walks if an entity of type same as that of the starting entity is encountered at depth $k < d$.

\subsubsection{Unique Entity Types (UET)}\label{sssec:uet} \hfill \break
In this scheme, we consider walks that have no two entities with the same type appearing in a single graph walk, making it the most restrictive pruning scheme out of the four.


\section{Evaluation}
\label{sec:evaluation}
We evaluate our approach on four different criteria. First, we analyze the  behavior of specificity computed for most relevant semantic relationships up to depth 4. Second, we study sensitivity of specificity computations on the parameters $N_{walks}$ and $|S|$ (from Algorithm \ref{alg:propSel}). Third, we evaluate the computation time and size of the extracted subgraphs by our proposed specificity-based biased random walk schemes against the baselines. Fourth, we evaluate our approach on the task of entity recommendation.

\subsection{Datasets}
We evaluate our approach on DBpedia which is one of the largest RDF repositories publicly available \cite{LehmannIJJKMHMK15}. We use the English version of DBpedia dataset from 2016-04\footnote{\url{http://wiki.dbpedia.org/dbpedia-version-2016-04}}. 
We create graph embedding for 3,000 entities of type \url{http://dbpedia.org/ontology/Film} (\textit{dbo:Film} for short)\footnote{\label{ft:link}Results based on \textit{dbo:Film} and more entity types can be found at \url{https://github.com/mrizwansaeed/Specificity}}.

\subsection{Experimental Setup and Results}
\label{sec:setup}
The first step is to compute specificity to find the set of most relevant semantic relationships with respect to entities of type \textit{dbo:Film}. The two main parameters used in the algorithm for computing specificity are size of seed set $S$ and number of bidirectional walks $N_{walks}$. We set these two parameters to 300 and 2000 respectively. These values come from the experiments in which we choose different values of the parameters to study their effect on specificity computation (Section \ref{sss:ndcg}). We randomly sample semantic relationships originating from entities in $S$ and select top-$25d$  semantic relationships based on frequency of occurrence, for each depth $d$. For $d > 1$, we use the most specific semantic relationships from depth $d-1$ to get a list of semantic relationships of depth $d$ for computing specificity\footnote{This is for populating $Q_{paths}$ in Algorithm \ref{alg:propSel}.}. We only consider those semantic relationships as relevant that have specificity greater than 50\%, which are then used for creating biased random walks for subgraph extraction. Data and models used to generate the results are available online\footnote{See footnote \ref{ft:link}.}.

\newpage
\subsubsection{Specificity as a Metric for Measuring Relevance} \label{sec:specificity_behavior} \hfill \break
\begin{figure*}[!ht]
\captionsetup[subfigure]{justification=centering}
\centering
\subfloat[Frequency of top-25 semantic relationships]
{\includegraphics[width=0.4\textwidth]{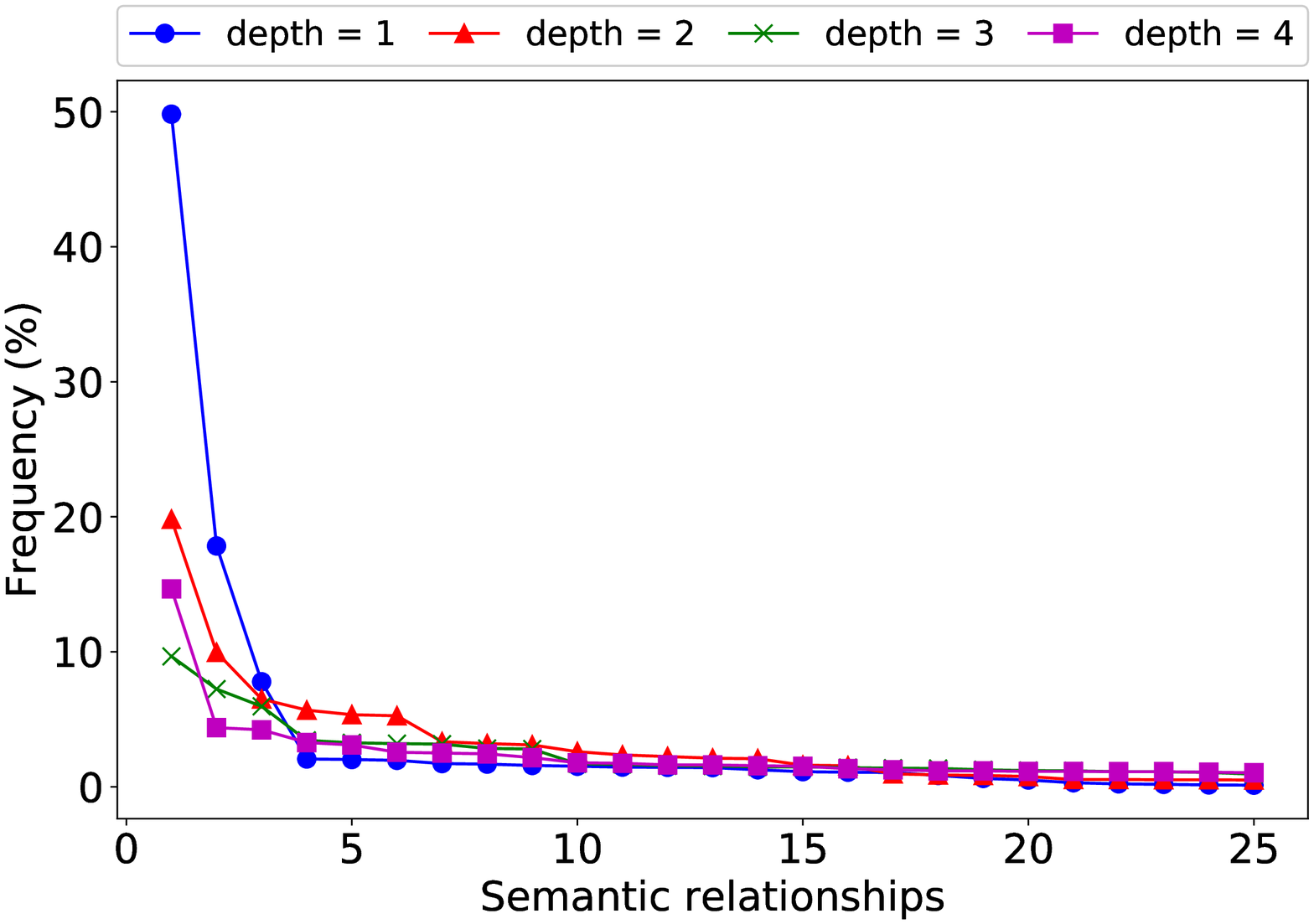}
\label{fig:frequency}}\quad
\subfloat[Specificity of top-25 semantic relationships]{\includegraphics[width=0.4\textwidth]{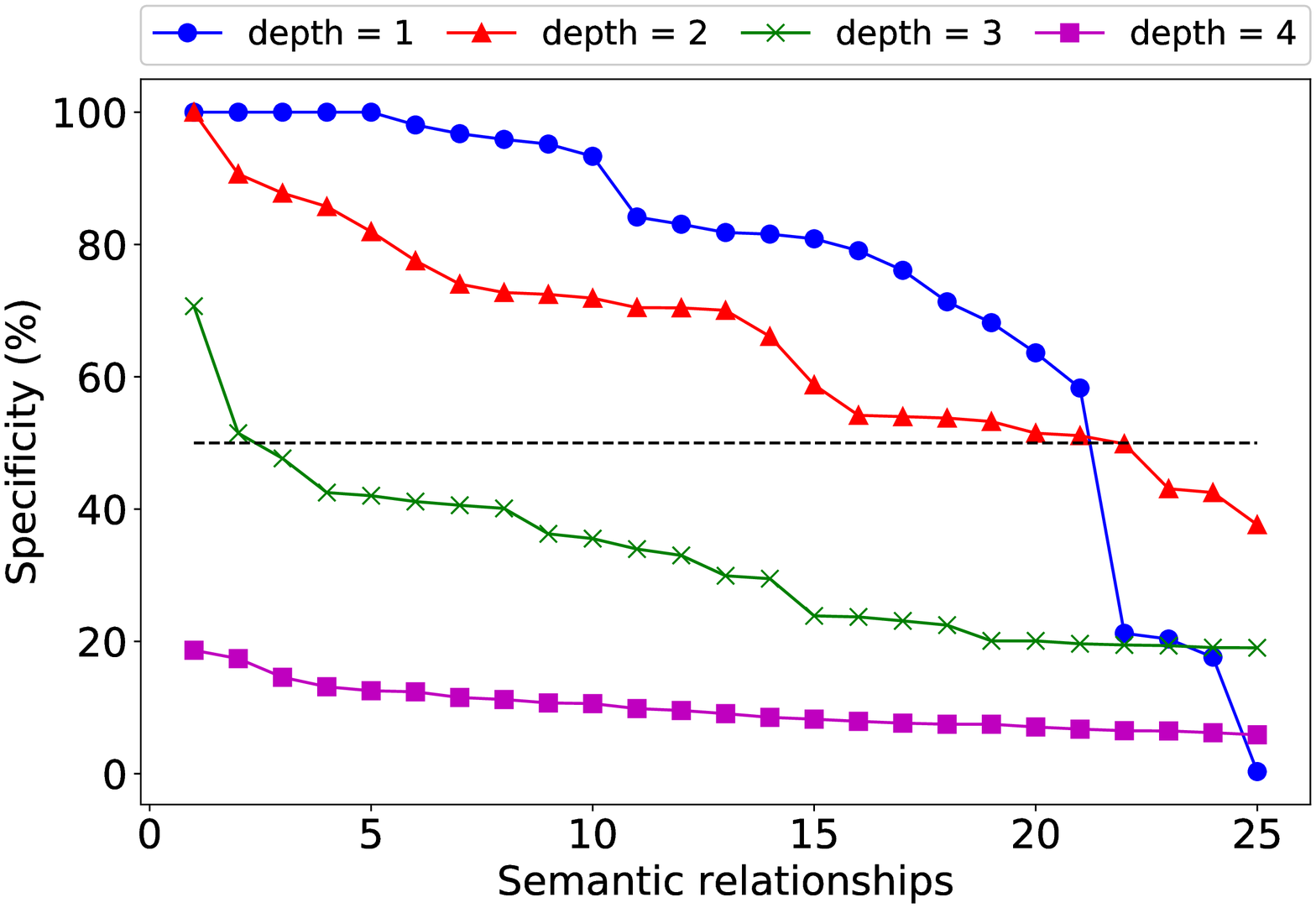}
\label{fig:specificity}}
\caption{Comparison of frequency and specificity}
\label{fig:metriccomparison}
\end{figure*}

Figure \ref{fig:metriccomparison} shows the top 25 semantic relationships sorted by their frequency and specificity for depths up to 4. Figure \ref{fig:frequency}, resembling power law curves, shows long tailed behavior for frequency of top occurring semantic relationships. As depth increases, frequency exhibits a flattened behavior due to rapid increase in number of possible semantic relationships at each depth $d$. Specificity of top-25 semantic relationships is shown in Figure \ref{fig:specificity} with the threshold of specificity drawn at 50\%. The specificity of a semantic relationship, here, is the probability of reaching any node of type \textit{dbo:Film} from a set of nodes ($V_{S,P^d}$ in Definition \ref{def:specificity}) by reverse walks in $G$ (or forward walks in $G^r$) using any arbitrary path of length $d$. This suggests that as $d$ increases, specificity of semantic relationships of length $d$ decreases. 
This can be seen in Figure \ref{fig:specificity} where, for every depth between 1 and 4, specificity is showing a more distinctive diminishing behavior as compared to frequency. 
This behavior is analogous to using a decaying factor, usually a function of depth \cite{PaulRMKS16}, that is used to assign low relevance to nodes farther away from target nodes. 
It can also be seen in Figure \ref{fig:specificity} that there are multiple instances of semantic relationships of length $d$ that have higher specificity than semantic relationships of lengths less than $d$.  
This indicates that specificity-based biased random walks allow both shallow (breadth-first) and deep (depth-first) exploration of the relevant neighborhoods around target entities. 
The ability to incorporate decaying behavior and interpolate between shallow and deep exploration allows specificity to quantify relevance of semantic relationships, of different lengths, at a more fine-grained level. 

\begin{table}
\centering
\captionsetup{justification=centering}
\begin{tabularx}{\columnwidth}{|p{4.2cm}|X|X|X|} \hline
\textbf{Semantic Relationships} & \textbf{Specificity} & \textbf{PageRank}\cite{Thalhammer2016} & \textbf{Frequency}\\
\hline
dbo:director,dbo:knownFor & 59.14 & 6.2 & 345\\
dbo:director,dct:subject & 1.05 & 823.53 & 73752\\
dbo:director,dbo:birthPlace & 0.03 & 200.33 & 7087\\
\hline    
\end{tabularx}
\caption{Comparison of relevance metrics for Example \ref{ex:batman} (Section \ref{ss:specificity})}
\label{table:specificitycomp}
\end{table}
Table \ref{table:specificitycomp} shows computed relevance of the three semantic relationships from Example \ref{ex:batman} (Section \ref{ss:specificity}) based on their specificity, PageRank, and frequency. The given PageRank values in column 3 are the average of non-normalized PageRank scores \cite{Thalhammer2016} of top-25 nodes linked to entities of type \textit{dbo:Film} by corresponding semantic relationship. The values of frequency, in last column, represent the number of occurrences of corresponding semantic relationship in DBpedia dataset. We argued that the semantic relationship \textit{dbo:director,dbo:knownFor} is more relevant to a film as compared to the other two. It can be seen from Table \ref{table:specificitycomp} that the proposed specificity based relevance metric is closer to our intuition as compared to other metrics. 

\subsubsection{Sensitivity of Specificity to $N_{walks}$ and $|S|$}\label{sss:ndcg}\hfill \break
\vspace{-0.6cm}
\begin{figure*}[!ht]
\captionsetup[subfigure]{justification=centering}
\centering
\subfloat[NDCG]{\includegraphics[width=0.48\textwidth]{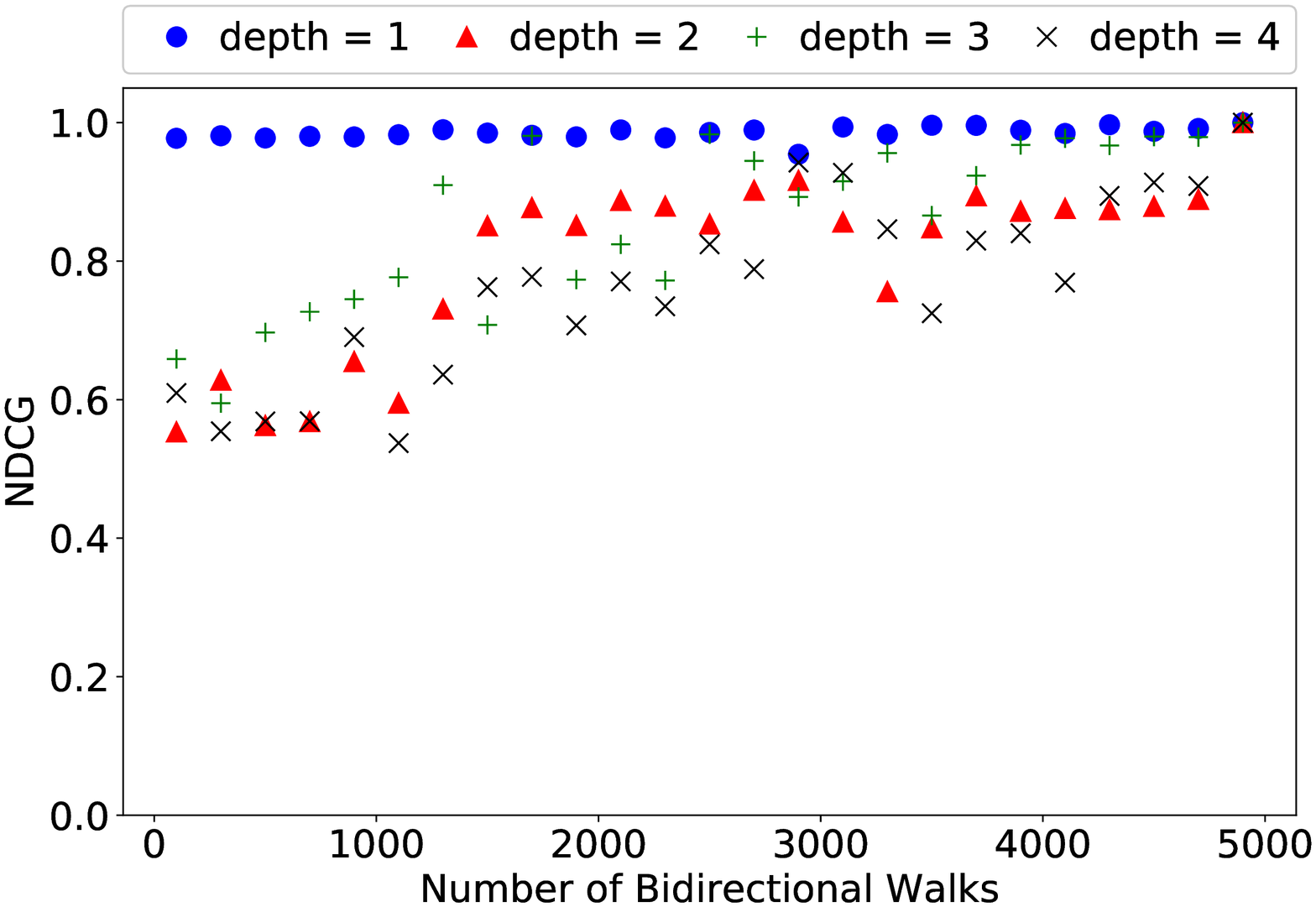}%
\label{fig:reversewalkeffectndcg}}
\hspace{0.01cm}
\subfloat[Computation time]
{\includegraphics[width=0.48\textwidth]{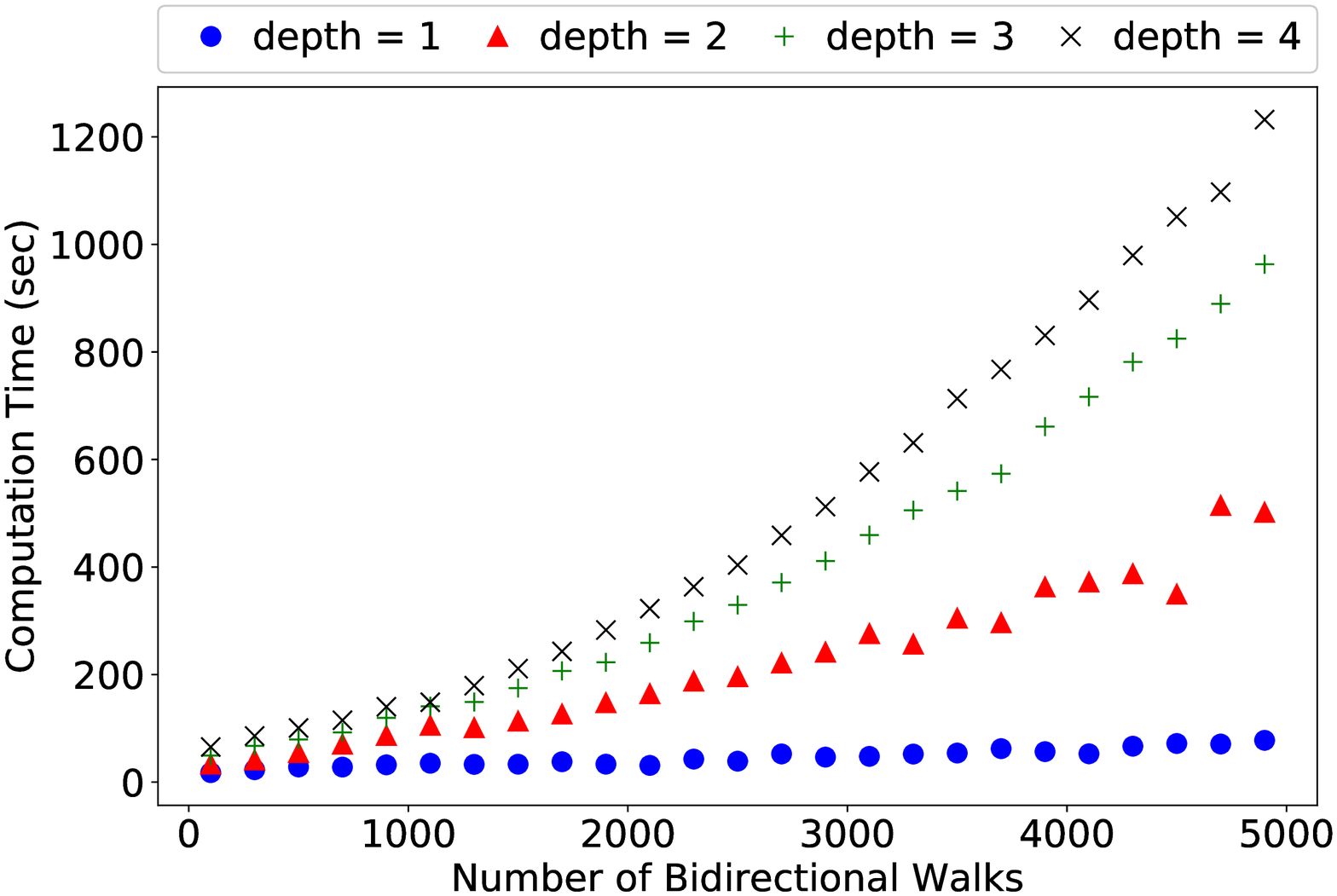}%
\label{fig:reversewalkeffectcomp}}
\caption{Effect of $N_{walks}$ on specificity}
\label{fig:reversewalkeffect}
\end{figure*}
\vspace{-0.5cm}

The algorithm for computing specificity uses bidirectional random walks, governed by two parameters: number of bidirectional walks ($N_{walks}$) and size of seed set $S$. To understand the effect of these parameters on specificity, we make the assumption that larger values for both parameters lead to better approximation of specificity because of inclusion of more nodes in the computations. 
We first compute specificity for a range of values for both parameters  and then take the computations performed using the largest values of $N_{walks}$ and $|S|$ as ground truth, instead of manually generated list of semantic relationships. To compare the different lists of semantic relationships sorted by specificity, we use \textit{NDCG} (Normalized Discounted Cumulative Gain) \cite{WangWLHL13}. 

Figure \ref{fig:reversewalkeffectndcg} shows the trend of NDCG values as $N_{walks}$ varies between 100 and 5000. For depth = 1, steady values of NDCG indicate that a few hundred bidirectional walks are sufficient for finding most relevant semantic relationships. 
For depths $\geq$ 2, there is an increasing but fluctuating trend of NDCG for larger values of $N_{walks}$, which becomes more pronounced for depth $\geq$ 3. We believe that the fluctuations come from the fact that as we go further away from the seed set $S$ during forward walks (Algorithm \ref{alg:reverse}), the number of possible paths for reverse walks increases substantially. Addressing this requires a larger value of $N_{walks}$ that allows the algorithm to traverse sufficiently large number of these reverse paths to provide a stable estimation of specificity. However, before selecting such a high value, it should be noted that the execution time of Algorithm \ref{alg:reverse}  
is directly affected by parameter $N_{walks}$. 
Figure \ref{fig:reversewalkeffectcomp} shows the computation times for specificity for different values of $N_{walks}$ and depth $d$. For $N_{walks} \approx 2000$, which gives $NDCG \geq 0.8$ for depths 1 and 2, it takes approximately 200 seconds to find semantic relationships of high specificity. For depth = 4 and $N_{walks} = 5000$, the computation of specificity takes approximately 1200 seconds. 
However, this computation is only needed to be performed once for each type of target entities in a KG. 

\begin{figure*}[!t]
\captionsetup[subfigure]{justification=centering}
\centering
\includegraphics[width=0.47\textwidth]{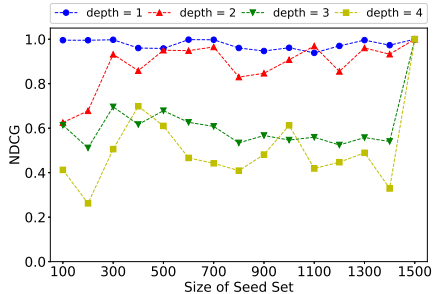}
\caption{Effect of $S$ on specificity computations}
\label{fig:seedsetsizeeffect}
\end{figure*}

Figure \ref{fig:seedsetsizeeffect} shows the behavior of changing the size of seed set $S$ on NDCG, for $N_{walks} = 1500$. It can be observed that for $|S| \geq$ 300, further increasing the size of seed set does not have significant effect on the order of sorted semantic relationships. This was expected, since the running time of Algorithm \ref{alg:reverse} only depends on the number of semantic relationships in input list $Q$ (which controls the outer loop, lines 3-14) and $N_{walks}$ (which controls the inner loop, lines 5-12). 

\subsubsection{Comparison of Size and Computation Time for Extracted Subgraphs}\hfill \break
We have chosen RDF2Vec \cite{RistoskiP16} (denoted as $R2V$ in the plots) as one of our baselines. The other baseline, we use, is PageRank-based biased random walks, which we denote as $R2V_{PR}$\footnote{We use the DBpedia PageRank \cite{Thalhammer2016} dataset from \url{http://people.aifb.kit.edu/ath/#DBpedia_PageRank}.}. We use $R2V_{Sp}$ to denote specificity-based biased random walks without pruning. The specificity-based pruned biased random walks are denoted as $NRSE_{Sp}$, $UE_{Sp}$, $NRST_{Sp}$, and $UET_{Sp}$. For the RDF data set, we generate 500 walks (based on \cite{RistoskiP16}) with depths 1, 2, and 3. 

\begin{figure*}[!t]
\captionsetup[subfigure]{justification=centering}
\centering
\subfloat[Size of extracted subgraph (in terms of number of walks)]{\includegraphics[width=0.45\textwidth]{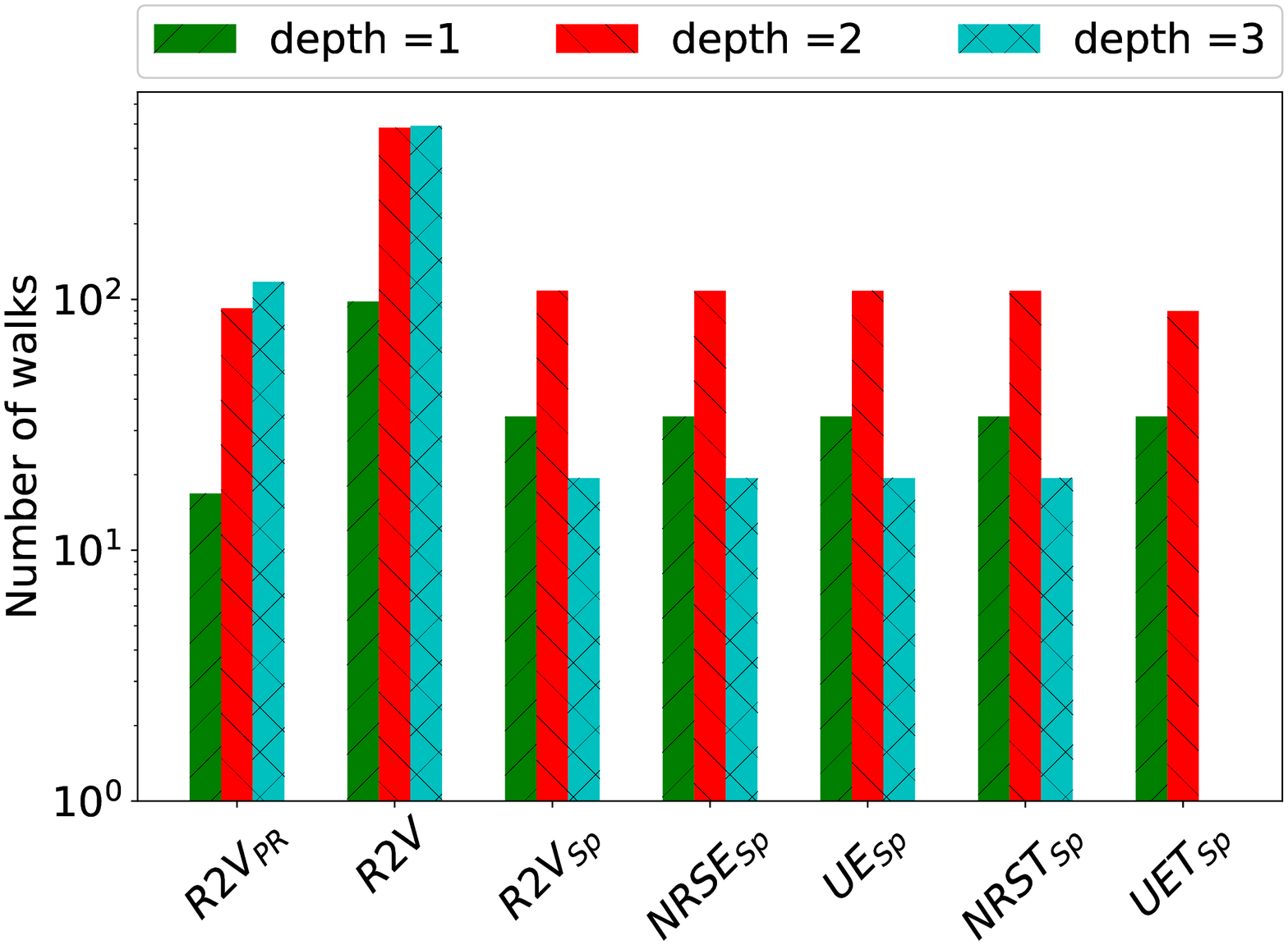}%
\label{fig:case_avgtokens}}\,
\subfloat[Average subgraph extraction time per entity]
{\includegraphics[width=0.45\textwidth]{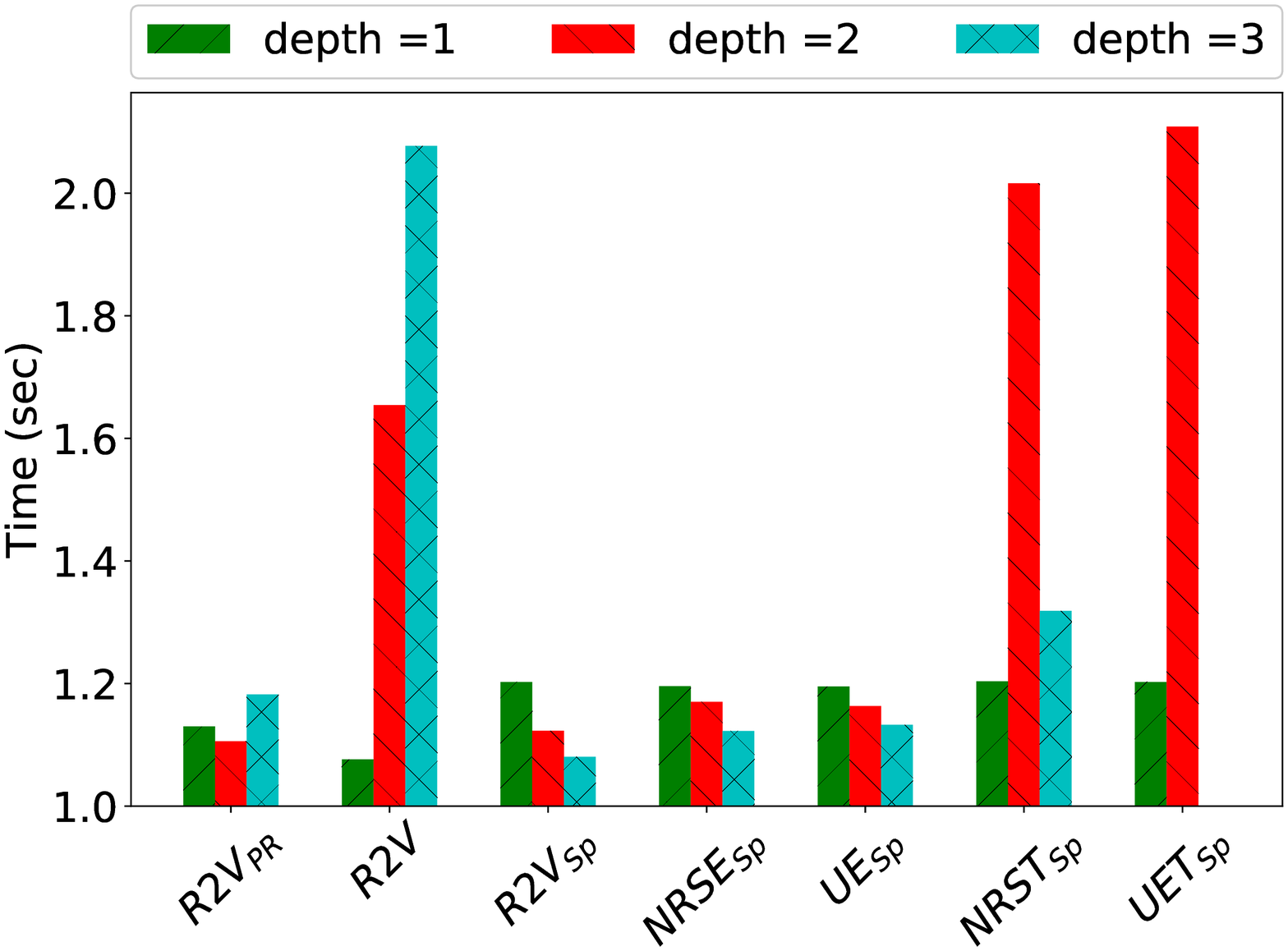}%
\label{fig:case_time}}
\caption{Comparison based on size of extracted subgraph and computation time}
\label{fig:schemescomparison}
\end{figure*}

Figure \ref{fig:case_avgtokens} shows that biased random walks are able to extract the representative entity subgraphs with fewer number of walks. This is mainly because of the reason that biased random walks extract subgraphs using a specific extraction template based on semantic relationships with higher specificity. This enables collection of fewer but more relevant nodes and edges than the unbiased random walks. For depth = 1, $R2V_{PR}$ generates the least number of walks. The DBpedia PageRank dataset \cite{Thalhammer2016} only provides PageRank values for entities with URIs. Therefore, $R2V_{PR}$ explores the KG only through object properties, excluding datatype properties and literals, resulting in fewer number of walks. All biased walks also take fewer time for subgraph extraction, except for $NRST_{Sp}$ and $UET_{Sp}$ because of the extensive checks needed to perform these walks (Section \ref{sec:prunedRW}). The most restrictive scheme $UET_{Sp}$ did not generate any walks for depth = 3.

\subsubsection{Suitability for Entity Recommendation task}\hfill \break
\begin{figure*}[!t]
\captionsetup[subfigure]{justification=centering}
\centering
\subfloat[Star Wars (1977) @5]
{\includegraphics[width=0.32\textwidth]{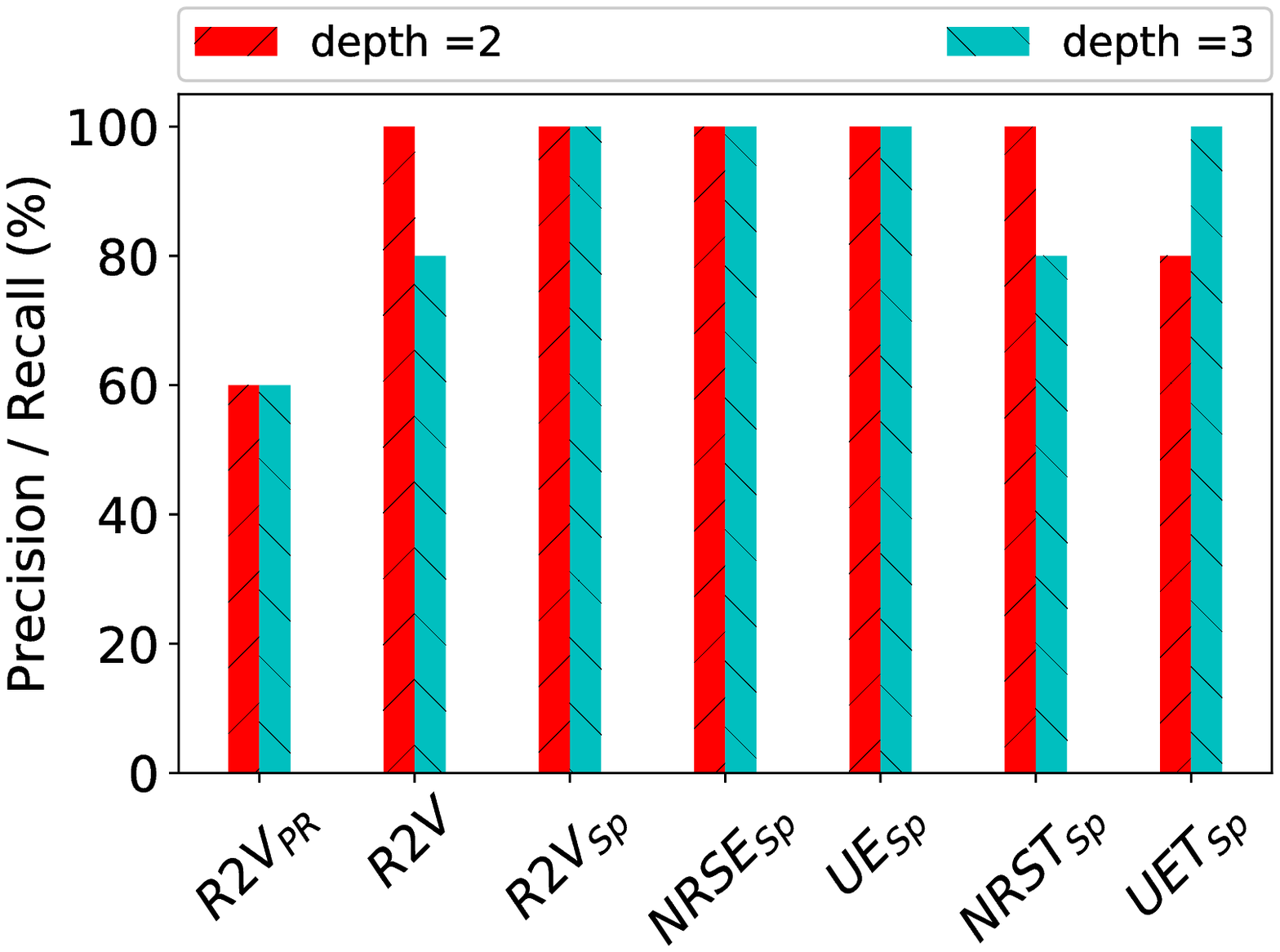}
\label{fig:starwars_accuracy}}
\subfloat[Star Trek: The Motion Picture (1979) @9]{\includegraphics[width=0.32\textwidth]{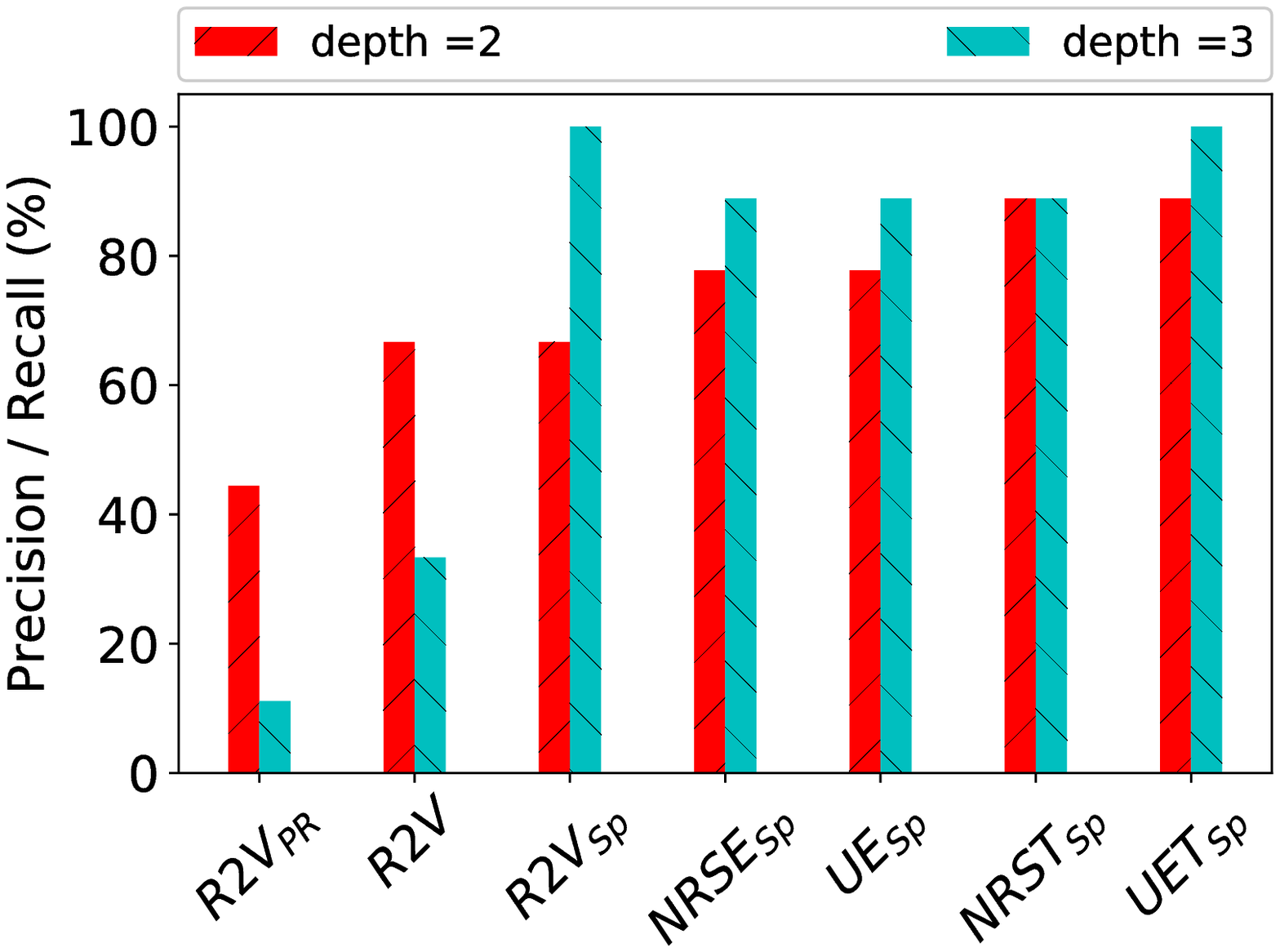}%
\label{fig:startrek_accuracy}}
\subfloat[Batman (1989) @5]
{\includegraphics[width=0.32\textwidth]{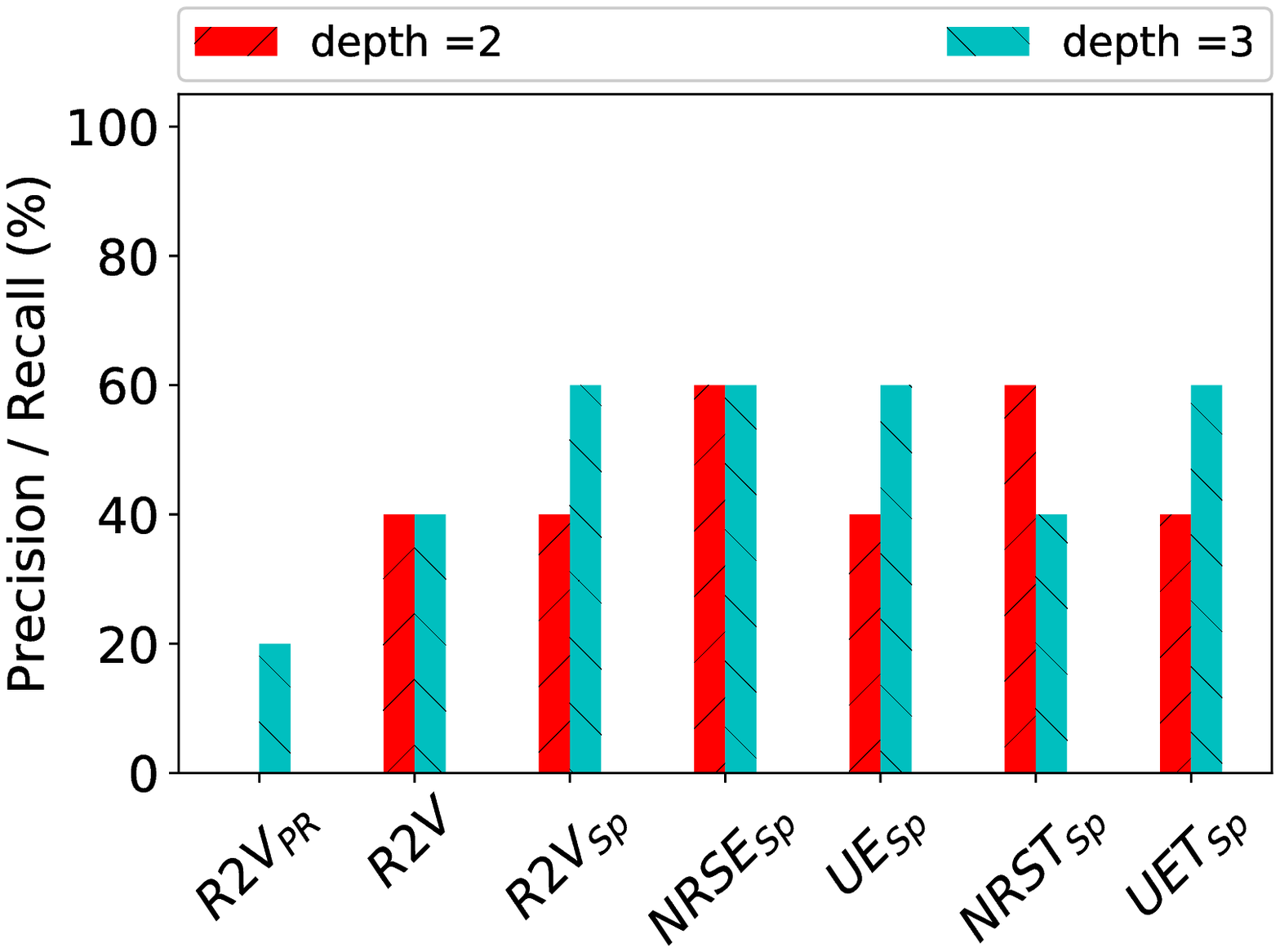}
\label{fig:batman_accuracy}}\\
\subfloat[LoTR: Fellowship of the Ring (2001) @3]{\includegraphics[width=0.32\textwidth]{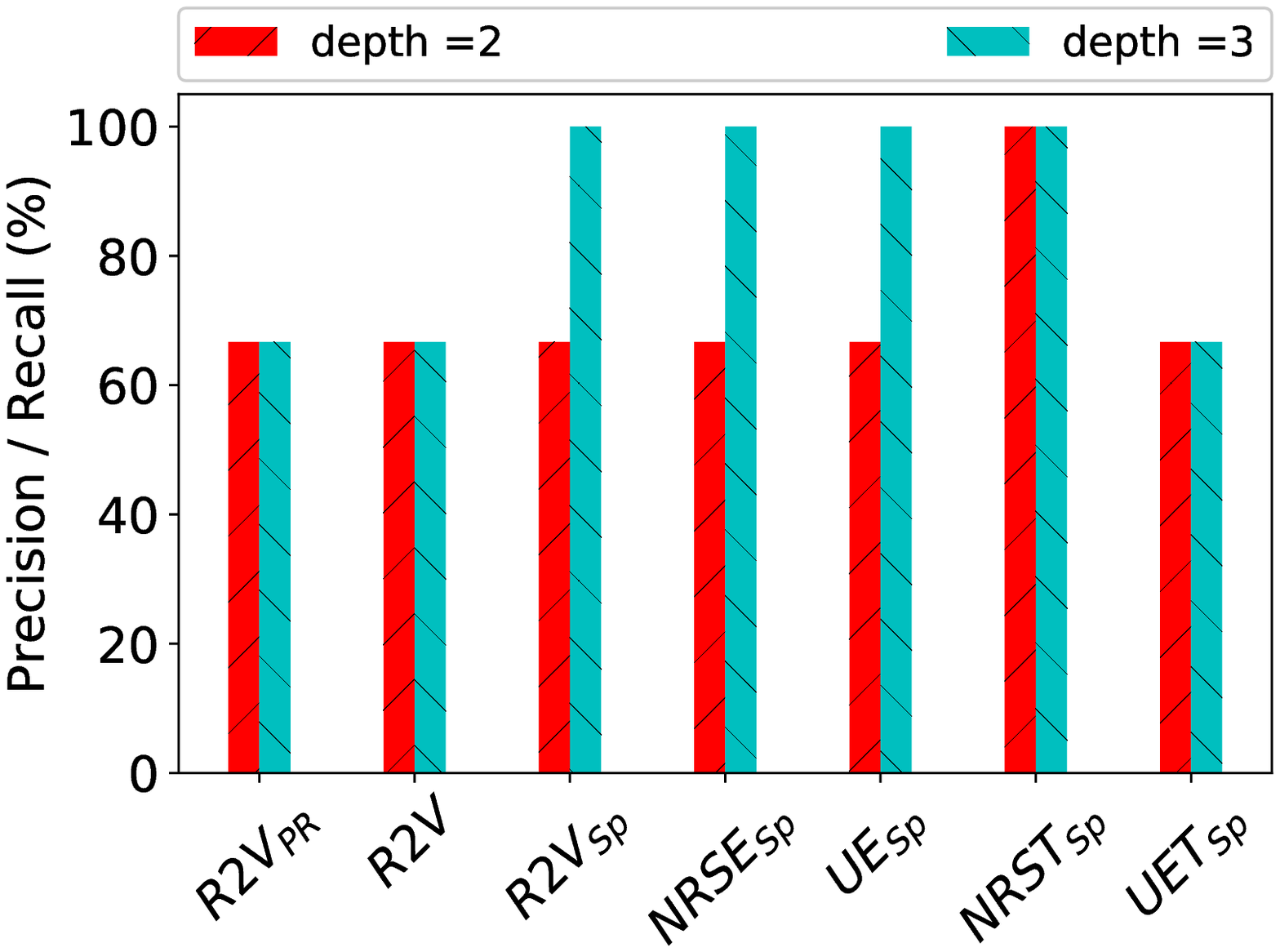}
\label{fig:harry_accuracy}}
\subfloat[King Kong (1933) @6]
{\includegraphics[width=0.32\textwidth]{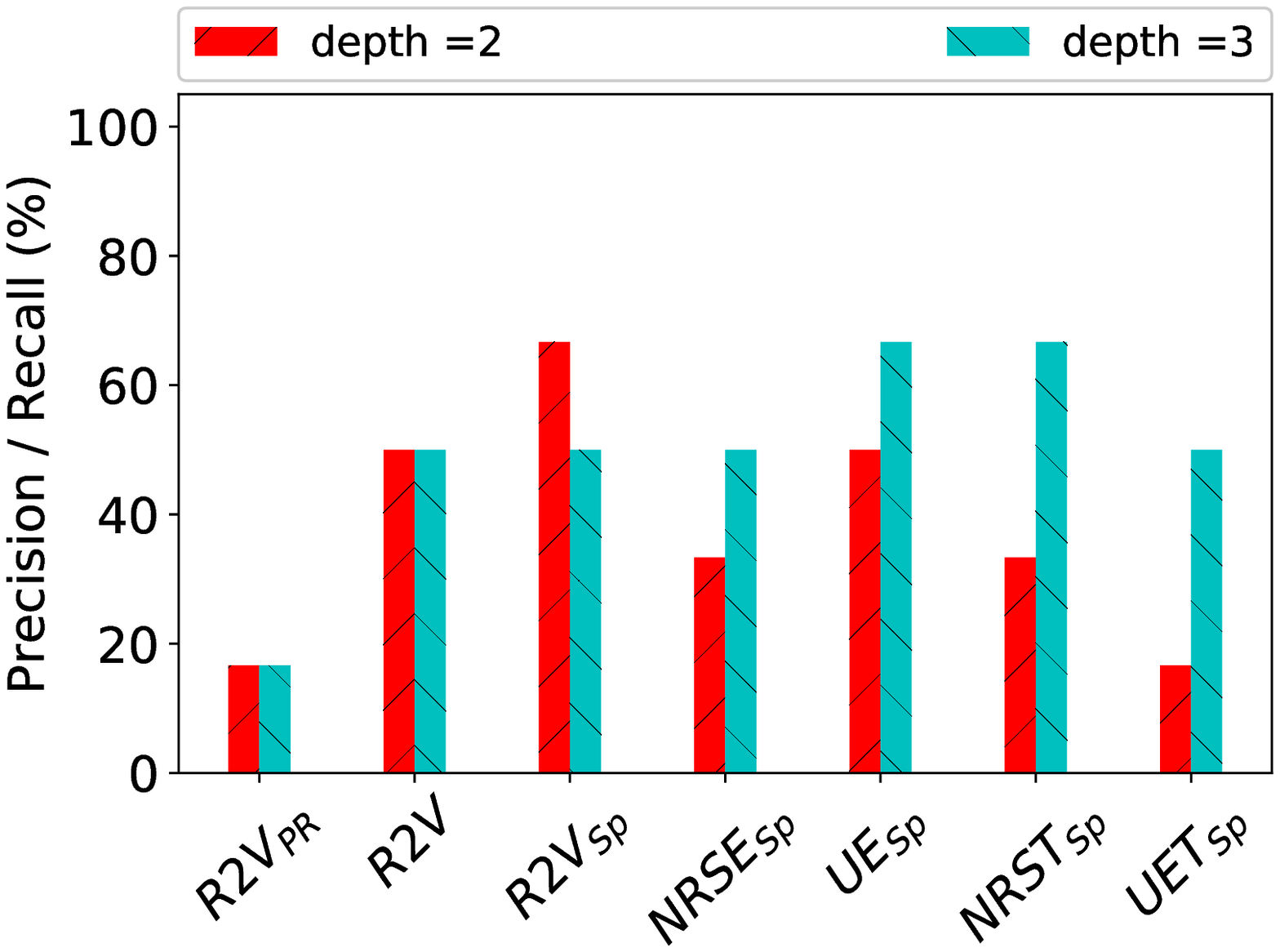}
\label{fig:kingkong_6_accuracy}}
\subfloat[Godzilla: King of Monsters! (1956) @21]{\includegraphics[width=0.32\textwidth]{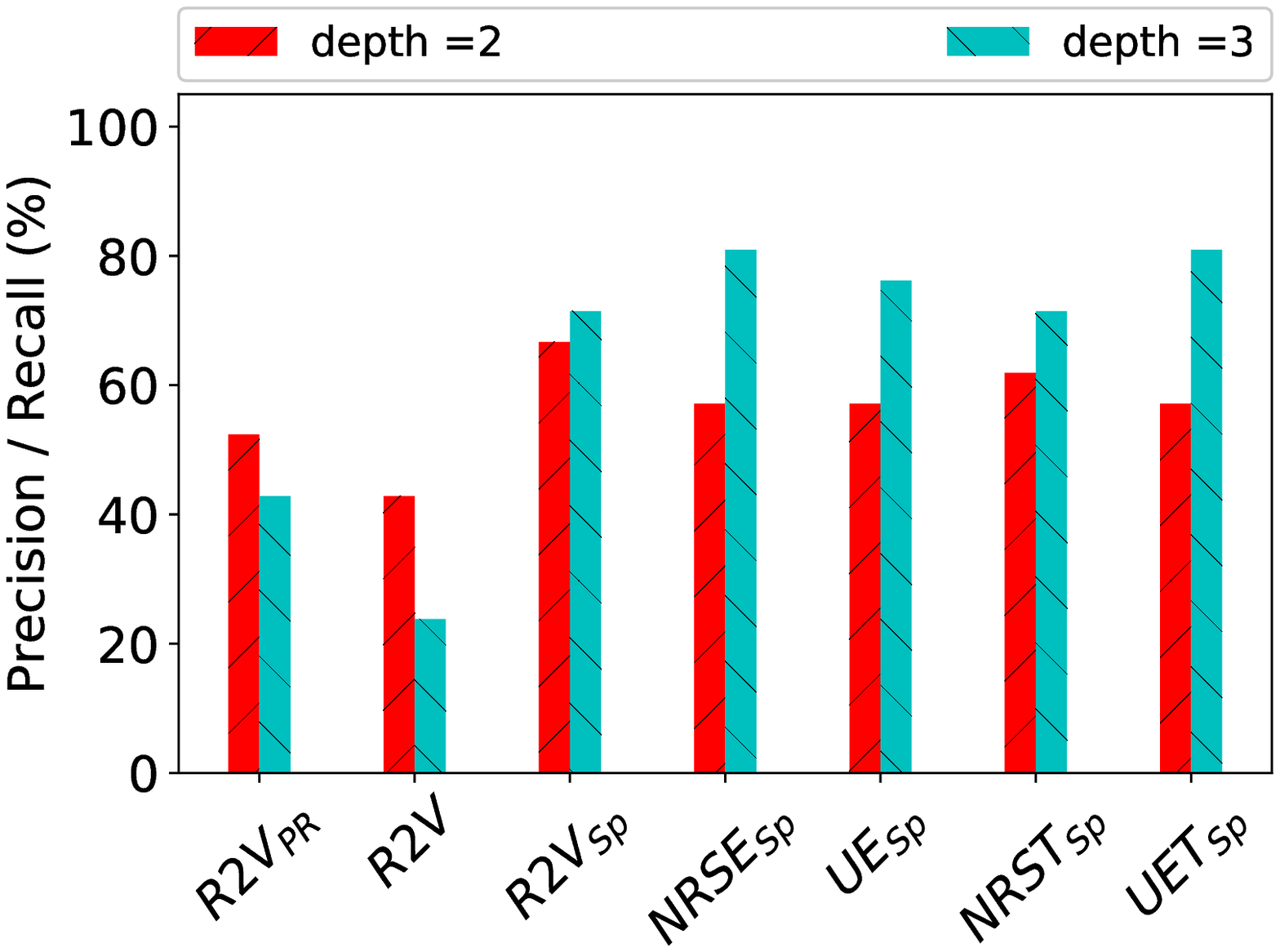}
\label{fig:godzilla_accuracy}}
\caption{Results for content-based recommendation task}
\label{fig:movieaccuracy}
\end{figure*}

We have shown that the specificity-based biased random walks extract more compact entity representations as compared to unbiased random walks. However, to prove that the compactness of extracted subgraphs is not a disadvantage, we use the graph embedding generated from the extracted substructures for the task of entity recommendation. Using the extracted subgraphs (represented as sequence of labels), we train Skip-gram models using the following parameters: dimensions of generated vectors = 500, window size = 10, negative samples = 25, iterations = 5 for each scheme and depth. All models\footnote{The trained models are available at \url{https://github.com/mrizwansaeed/Specificity}} for depth $d > 1$ are trained using sequences generated for both depths 1 and $d$. The parameters for this experiment are based on RDF2Vec \cite{RistoskiP16}. 

Figure \ref{fig:movieaccuracy} shows results for six different movies selected for entity recommendation task. The search key for each recommendation task is provided as the caption of the corresponding plot. We have selected movies that are part of movie franchises or series. The ground truth, then simply, consists of other movies in their respective franchises or movie series. For evaluation, we retrieve \textit{top-k} similar movies using the trained word2vec models. $k$ is chosen to be the total number of movies in the ground truth excluding the search key, which makes precision and recall to be the same. Note, that \textit{The Lord of Rings: Fellowship of the Ring (2001)} has two sequels but the value of $k$ given in caption is 3. This is because of the entity $dbr:The\_Lord\_of\_the\_Rings\_(film\_series)$ of type \textit{dbo:Film} in the DBpedia dataset which corresponds to the Wikipedia page about the entire film series\footnote{\url{https://en.wikipedia.org/wiki/The_Lord_of_the_Rings_(film_series)}}. 
The results show that specificity-based schemes in majority of cases perform better than the baselines. Even, where our proposed strategies have comparable results (Figure \ref{fig:starwars_accuracy}), it must be noted that this accuracy was achieved using more compact entity representations as compared to $R2V$ (Figure \ref{fig:case_avgtokens}). 
 
\section{Conclusion}
\label{sec:conclusion}
Graph embedding is an effective method of preparing KGs for AI and ML techniques. However, to generate appropriate representations, it is imperative to identify the most relevant representative subgraphs for target entities. In this paper, we presented specificity as a useful metric for finding the most relevant semantic relationships for target entities of a given type. Our bidirectional random walks-based approach for computing specificity is suitable for large scale KGs of any structure and size. We have shown, through experimental evaluation, that the metric of specificity incorporates a fine-grained decaying behavior for semantic relationships. It has the inherent ability to interpolate between the extreme exploration strategies: BFS and DFS. We used specificity-based biased random walks to extract compact representations of target entities for generating graph embedding. These generated representations have better performance with respect to baseline approaches when used for our selected task of entity recommendation. For future work, we will study the effects on other tasks in which specificity-based graph embedding can be used, such as link predictions and classifications in KGs. 

\section*{Acknowledgment}
This work is supported by Chevron Corp. under the joint project, Center for Interactive Smart Oilfield Technologies (CiSoft), at the University of Southern California.

\bibliographystyle{splncs03}
\bibliography{iswc}

\begin{thebibliography}{10}
\providecommand{\url}[1]{\texttt{#1}}
\providecommand{\urlprefix}{URL }

\bibitem{AggarwalAZB15}
Aggarwal, N., Asooja, K., Ziad, H., Buitelaar, P.: Who are the american vegans
  related to brad pitt?: Exploring related entities. In: Proceedings of the
  24th International Conference on World Wide Web Companion, {WWW} 2015

\bibitem{AtzoriD14}
Atzori, M., Dessi, A.: Ranking dbpedia properties. In: 2014 {IEEE} 23rd
  International {WETICE} Conference, {WETICE} 2014

\bibitem{berners2001semantic}
Berners-Lee, T., Hendler, J., Lassila, O.: The semantic web. Scientific
  american  284(5),  34--43 (2001)

\bibitem{bizer2009linked}
Bizer, C., Heath, T., Berners-Lee, T.: Linked data-the story so far.
  International journal on semantic web and information systems  5(3),  1--22
  (2009)

\bibitem{CochezRPP17}
Cochez, M., Ristoski, P., Ponzetto, S.P., Paulheim, H.: Biased graph walks for
  {RDF} graph embeddings. In: Proceedings of the 7th International Conference
  on Web Intelligence, Mining and Semantics, {WIMS} 2017

\bibitem{GabrilovichM07}
Gabrilovich, E., Markovitch, S.: Computing semantic relatedness using
  wikipedia-based explicit semantic analysis. In: {IJCAI} 2007, Proceedings of
  the 20th International Joint Conference on Artificial Intelligence, 2007

\bibitem{GroverL16}
Grover, A., Leskovec, J.: node2vec: Scalable feature learning for networks. In:
  Proceedings of the 22nd {ACM} {SIGKDD} International Conference on Knowledge
  Discovery and Data Mining 2016

\bibitem{LealRQ12}
Leal, J.P., Rodrigues, V., Queir{\'{o}}s, R.: Computing semantic relatedness
  using dbpedia. In: 1st Symposium on Languages, Applications and Technologies,
  {SLATE} 2012

\bibitem{LehmannIJJKMHMK15}
Lehmann, J., Isele, R., Jakob, M., Jentzsch, A., Kontokostas, D., Mendes, P.N.,
  Hellmann, S., Morsey, M., van Kleef, P., Auer, S., Bizer, C.: Dbpedia - {A}
  large-scale, multilingual knowledge base extracted from wikipedia. Semantic
  Web  6(2),  167--195 (2015)

\bibitem{LoschBR12}
L{\"{o}}sch, U., Bloehdorn, S., Rettinger, A.: Graph kernels for {RDF} data.
  In: The Semantic Web: Research and Applications - 9th Extended Semantic Web
  Conference, {ESWC} 2012

\bibitem{Mikabs-1301-3781}
Mikolov, T., Chen, K., Corrado, G., Dean, J.: Efficient estimation of word
  representations in vector space. CoRR  abs/1301.3781 (2013)

\bibitem{MikolovSCCD13}
Mikolov, T., Sutskever, I., Chen, K., Corrado, G.S., Dean, J.: Distributed
  representations of words and phrases and their compositionality. In: 27th
  Annual Conference on Neural Information Processing Systems - {NIPS} 2013

\bibitem{nickel2016review}
Nickel, M., Murphy, K., Tresp, V., Gabrilovich, E.: A review of relational
  machine learning for knowledge graphs. Proceedings of the IEEE  104(1),
  11--33 (2016)

\bibitem{PaulRMKS16}
Paul, C., Rettinger, A., Mogadala, A., Knoblock, C.A., Szekely, P.A.: Efficient
  graph-based document similarity. In: The Semantic Web. Latest Advances and
  New Domains - 13th International Conference, {ESWC} 2016

\bibitem{PenningtonSM14}
Pennington, J., Socher, R., Manning, C.D.: Glove: Global vectors for word
  representation. In: Proceedings of the 2014 Conference on Empirical Methods
  in Natural Language Processing, {EMNLP} 2014

\bibitem{PerozziAS14}
Perozzi, B., Al{-}Rfou, R., Skiena, S.: Deepwalk: online learning of social
  representations. In: The 20th {ACM} {SIGKDD} International Conference on
  Knowledge Discovery and Data Mining, {KDD} 2014

\bibitem{rettinger2012mining}
Rettinger, A., L{\"o}sch, U., Tresp, V., d'Amato, C., Fanizzi, N.: Mining the
  semantic web. Data Mining and Knowledge Discovery  24(3),  613--662 (2012)

\bibitem{RistoskiP16}
Ristoski, P., Paulheim, H.: Rdf2vec: {RDF} graph embeddings for data mining.
  In: The Semantic Web - {ISWC} 2016

\bibitem{ShervashidzeVPMB09}
Shervashidze, N., Vishwanathan, S.V.N., Petri, T., Mehlhorn, K., Borgwardt,
  K.M.: Efficient graphlet kernels for large graph comparison. In: Proceedings
  of the Twelfth International Conference on Artificial Intelligence and
  Statistics, {AISTATS} 2009

\bibitem{SunHYYW11}
Sun, Y., Han, J., Yan, X., Yu, P.S., Wu, T.: Pathsim: Meta path-based top-k
  similarity search in heterogeneous information networks. {PVLDB}  4(11),
  992--1003 (2011)

\bibitem{Thalhammer2016}
Thalhammer, A., Rettinger, A.: {PageRank on Wikipedia: Towards General
  Importance Scores for Entities}. In: The Semantic Web: ESWC 2016 Satellite
  Events, 2016

\bibitem{TzitzikasLZ12}
Tzitzikas, Y., Lantzaki, C., Zeginis, D.: Blank node matching and {RDF/S}
  comparison functions. In: The Semantic Web - {ISWC} 2012

\bibitem{VriesR15}
de~Vries, G.K.D., de~Rooij, S.: Substructure counting graph kernels for machine
  learning from {RDF} data. J. Web Sem.  35,  71--84 (2015)

\bibitem{WangWLHL13}
Wang, Y., Wang, L., Li, Y., He, D., Liu, T.: A theoretical analysis of {NDCG}
  type ranking measures. In: {COLT} 2013 - The 26th Annual Conference on
  Learning Theory, 2013

\bibitem{YanardagV15}
Yanardag, P., Vishwanathan, S.V.N.: Deep graph kernels. In: Proceedings of the
  21th {ACM} {SIGKDD} International Conference on Knowledge Discovery and Data
  Mining, 

\end{thebibliography}

\end{document}